\documentclass[]{fairmeta}

\usepackage[utf8]{inputenc} 
\usepackage[T1]{fontenc}    
\usepackage{hyperref}       
\usepackage{url}            
\usepackage{booktabs}       
\usepackage{amsfonts}       
\usepackage{amsmath}
\usepackage{nicefrac}       
\usepackage{microtype}      
\usepackage{xcolor}         
\usepackage{enumitem}      
\usepackage{tabularx}
\usepackage{multicol,multirow}
\usepackage{colortbl}
\usepackage{tipa}
\usepackage{caption}
\usepackage{subcaption}
\usepackage{graphicx}
\usepackage{multirow}
\usepackage{rotating}
\usepackage[noabbrev]{cleveref} 
\usepackage{wrapfig}   

\newcolumntype{C}{>{\centering\arraybackslash}X}
\newcolumntype{R}{>{\raggedleft\arraybackslash}X}
\newcolumntype{L}{>{\raggedright\arraybackslash}X}

\definecolor{darkblue}{RGB}{14, 64, 123}
\hypersetup{%
    colorlinks,
    linkcolor=darkblue,
    citecolor=darkblue,
    urlcolor=darkblue,
}

\newcommand{\ourmodel}{AU-Net}
\newcommand{\dm}{data-to-model ratio}

\title{From Bytes to Ideas:\\ Language Modeling with Autoregressive U-Nets} 

\author[1,*]{Mathurin Videau}
\author[1,*]{Badr Youbi Idrissi}
\author[3]{Alessandro Leite}
\author[2]{Marc Schoenauer}
\author[1]{Olivier Teytaud}
\author[1]{David Lopez-Paz}

\affiliation[1]{FAIR at Meta}
\affiliation[2]{TAU, INRIA and LISN, CNRS \& Université Paris-Saclay}
\affiliation[3]{INSA Rouen Normandy, LITIS, Rouen, France}

\contribution[*]{Equal contribution}

\abstract{%
  Tokenization imposes a fixed granularity on the input text, freezing how a language model operates on data and how far in the future it predicts. 
  Byte Pair Encoding (BPE) and similar schemes split text once, build a static vocabulary, and leave the model stuck with that choice. 
  We relax this rigidity by introducing an autoregressive U-Net that learns to embed its own tokens as it trains. 
  The network reads raw bytes, pools them into words, then pairs of words, then up to 4 words, yielding a multi-scale representation of the sequence. 
  At deeper stages, the model must predict further into the future\textemdash{}anticipating the next few words rather than the next byte\textemdash{}so deeper stages focus on broader semantic patterns while earlier stages handle fine details. 
  When carefully tuning and controlling pretraining compute, shallow hierarchies are on par with strong BPE baselines, and deeper hierarchies exhibit a promising trend. 
  Because tokenization now lives inside the model, the same system can handle character-level tasks and carry knowledge across low-resource languages.
}

\date{\today}
\correspondence{Mathurin Videau at \email{mathurin.videau@gmail.com}}

\metadata[Code]{\url{https://github.com/facebookresearch/lingua/tree/main/apps/aunet}}

\begin{document}

\maketitle


\begin{figure}[ht]
\centering
\includegraphics[width=0.94\textwidth]{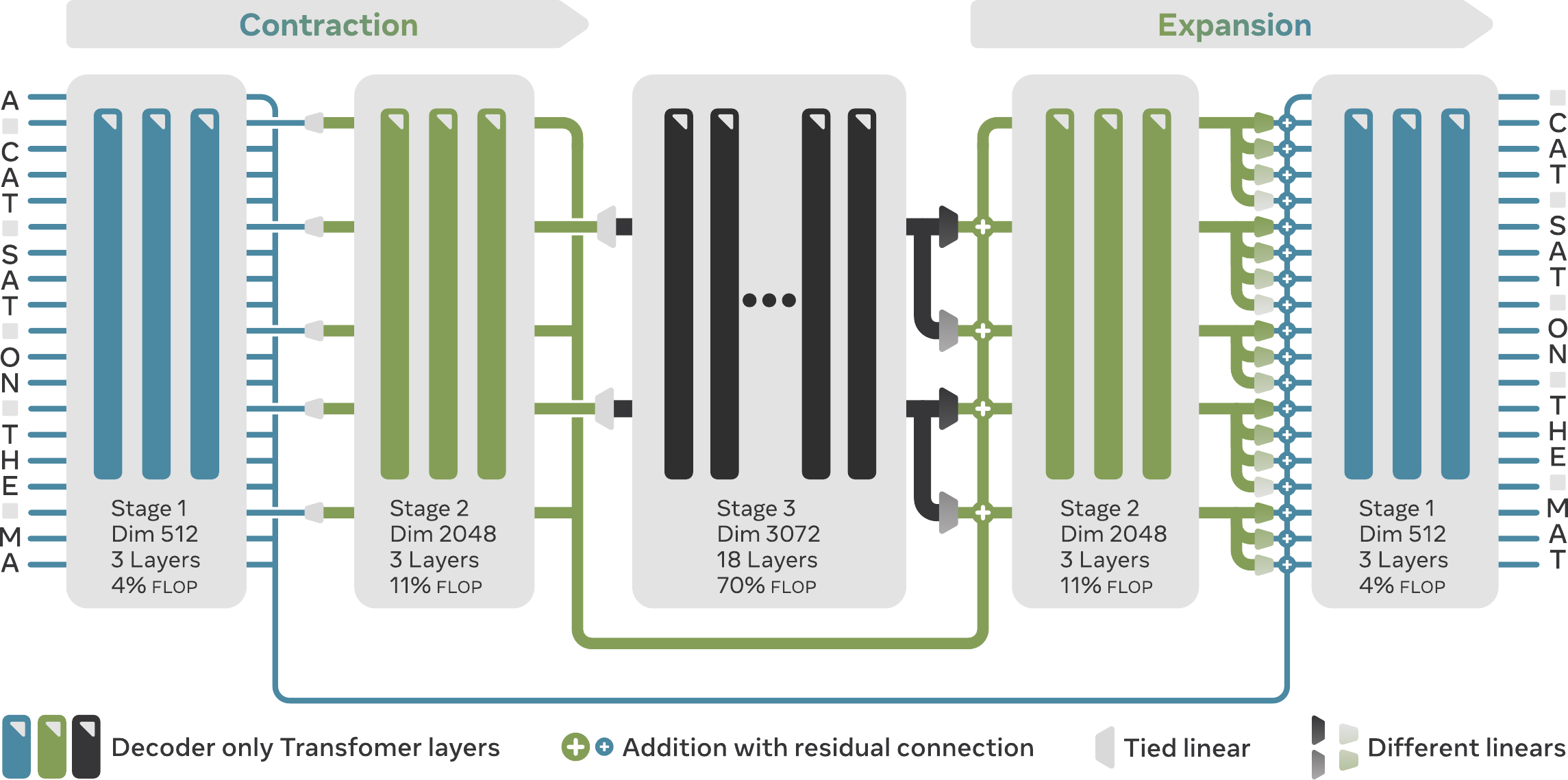}
\caption{\small Three-stage Autoregressive U-Net (\ourmodel{}).  
          The model executes from left to right. The contracting path compresses the sequence in two steps: Stage~1 processes raw bytes, Stage~2 keeps only the vector at each word boundary, and Stage~3 keeps one vector per two words. Each contraction and expansion step supports arbitrary pooling and upsampling patterns.
          After the deepest stage, the expanding path reverses the contracting path by duplicating each coarse vector and applying position-specific linear layers. 
          These are combined with skip connections from the contracting path, gradually restoring sequence length and blending in high-level information. 
          Deeper stages predict further ahead and capture broad semantics, while shallower stages refine local detail.}
\label{fig:main}
\end{figure}

\section{Introduction}\label{sec:intro}
Language models are about uncovering patterns in a sequence so they can guess what comes next.  
Before any of that happens, we must decide what the pieces of that sequence—the \emph{tokens}—actually are.  
That choice is usually frozen in advance by a \emph{tokeniser} that chops raw text into discrete units long before training begins.  
Consider the sentence ``The quick brown fox. ''  
A \emph{character}-level tokeniser feeds the model the stream \{\texttt{T, h, e, \textvisiblespace, q, u}\} and asks it to predict the next letter \texttt{i}.  
A \emph{word}-level tokeniser, in contrast, hands over \{\texttt{The}, \texttt{quick}\} and expects the model to guess \texttt{brown} in one shot.  
Finer cuts lead to larger sequences and shorten the look-ahead window, whereas coarser cuts lead to shorter sequences but make each token rarer and harder to compare and predict.  
Regardless of granularity, some form of tokenisation is unavoidable: a sequence must exist before any Transformer can run.


\textbf{Byte-Pair Encoding} (BPE) followed by a simple embedding table is by far the most popular approach. 
It works by repeatedly merging the most frequent byte sequences in the training text until a preset vocabulary limit is reached. 
This procedure leaves practitioners with just two intuitive \emph{dials}.  
The first dial is the \emph{training corpus}: whichever text one feeds the algorithm—English prose, source code, or a multilingual mix—determines which patterns are merged and therefore what the final tokens look like.  
The second dial is the \emph{vocabulary size}: raising this limit lets the merge process run for more steps, producing longer tokens and shorter sequences at the cost of a larger embedding table and output softmax.  

Most issues with tokenisation stem from the embedding operation rather than the splitting act itself. 
Each token is typically mapped to an independent vector, meaning the network sees only opaque identifiers and must rediscover, for instance, that \textit{strawberry} and \textit{strawberries} share nine letters. 
This reliance on isolated embeddings hampers symbol-level reasoning and complicates transfer to dialects or rare languages. 
Finally, this splitting is most often a preprocessing step, locking in a single level of granularity for all subsequent model layers (see \Cref{sec:splitting}).

To address these limits, our \textbf{Autoregressive U-Net} (\Cref{sec:ARUNet}), or \ourmodel{} (`oh-net', \textipa{/\'oU nEt/}), learns to embed information directly from raw bytes, and allows for multiple stages of splitting. 
The purpose of an embedding is to map tokens to vectors. 
Instead of using a lookup table, we use attention directly to embed the tokens.
Self-attention allows vectors at any position to summarize the entire preceding context. 
This enables a simple pooling mechanism: we select these contextualized vectors at word boundaries (\ourmodel{}-2), then word pairs (\ourmodel{}-3), and up to four-word chunks (\ourmodel{}-4), forming a multi-stage embedding hierarchy. 
This U-Net like architecture contracts sequences, preserving detail with skip connections, before expanding them. 
During expansion, vectors representing coarser information are injected back into more fine grained representations. 
Deeper stages, by operating on compressed views, inherently need to anticipate multiple words ahead, similar to multi-token prediction~\citep{mtp} but without auxiliary losses.
This effect allows deeper stages to guide shallower stages at the semantic level, while letting them handle finer details like spelling.

\textbf{Contributions} (quantified in \Cref{sec:experiments}).
\begin{enumerate}[label=\textbf{C\arabic*.}, leftmargin=2.4em]
    \item \textit{Adaptive multi-level hierarchy}.  
          We train up to four end-to-end embedding stages with arbitrary, user-specified split functions, extending prior work that relies either on fixed pooling or shallow hierarchies.
    \item \textit{Infinite vocab size}.
         By operating directly on bytes, our model avoids predefined vocabularies and memory-heavy embedding tables, allowing an unlimited number of unique tokens.
    \item \textit{Strong performance and scaling}.  
          Under identical pre-training budgets, a single level matches strong BPE baselines, and a two or three-level hierarchy shows promising scaling trends. A selection of the results are presented in Table~\ref{tab:datamodel}.
    \item \textit{Practical Efficiency} .
          We maintain comparable GPU throughput in wall-clock time instead of purely theoretical compute gains. Our code is available in Meta Lingua~(\cite{meta_lingua})\footnote{\url{https://github.com/facebookresearch/lingua/tree/main/apps/aunet}}.
    \item \textit{Stable scaling laws}.  
          We show that moving from token to byte-level training  demands new batch size and learning rate formulas to get smooth optimization.
\end{enumerate}


\section{Method}
\subsection{Autoregressive U-Net}\label{sec:ARUNet}

\begin{wraptable}{r}{0.44\linewidth}
  \centering
  \scriptsize
\caption{1B equivalent on 370B tokens}
\label{tab:datamodel-1b-370b}
\begin{tabularx}{\linewidth}{lcccc}
  \toprule
  Model & FLOP & Hellaswag & MMLU & GSM8k \\ \midrule
  BPE &4e21& 70.2 & 27.0 & 4.4 \\
  \ourmodel{}~2&3e21& 69.9 & 28.8 & 3.0 \\
  \ourmodel{}~3&4e21& 72.9 & 28.0 & 3.7 \\
  \ourmodel{}~4&5e21& \textbf{73.7} & \textbf{31.7} & \textbf{5.3} \\ \bottomrule
\end{tabularx}
\end{wraptable} 

Inspired by U-Net-like architectures~\citep{unet,hourglass}, we propose an autoregressive hierarchical model for language modeling, illustrated in~\cref{fig:main}. 
This architecture features a \emph{contracting path}, which compresses the input sequence, and an \emph{expanding path}, which reconstructs it. Both paths are fully \emph{adaptive}: they do not require fixed pooling or upsampling sizes.
Pooling and upsampling operations can be designed independently, even if we choose to make them symmetrical in this paper.
The only requirement is a \emph{splitting function}, which specifies the positions in the sequence where pooling should occur. This function is detailed in~\cref{sec:splitting}.

Our architecture is \emph{monolithic}: unlike recent approaches~\citep{blt,hierarchicalattention} that use local models, we apply attention globally at each stage (or within a sliding window), allowing every input to attend to previous inputs. 
This ensures that words or word groups are not processed in isolation. 
To preserve fine-grained information that might be lost during contraction, we introduce skip connections between stages, following the approach in~\citet{unet} and~\citet{hourglass}. 
We also increase the hidden dimension at each stage in proportion to its contraction factor, enabling richer representations as the sequence is contracted. 
To keep computation tractable at the byte-level stage (Stage 1), where sequences are longest, we restrict attention to a window.

\subsubsection{Pooling and Upsampling}\label{sec:poolnup}

Since our pooling and upsampling are adaptive, we cannot rely on fixed window sizes. 
To address this, we explored several pooling and upsampling strategies. 
In this section, we describe the method used in all experiments reported in the main text. 
A complete description of the alternatives and ablation results can be found in the appendix~\ref{app:ablation}.
\begin{figure}[ht]
  \centering
  \includegraphics[width=0.7\textwidth]{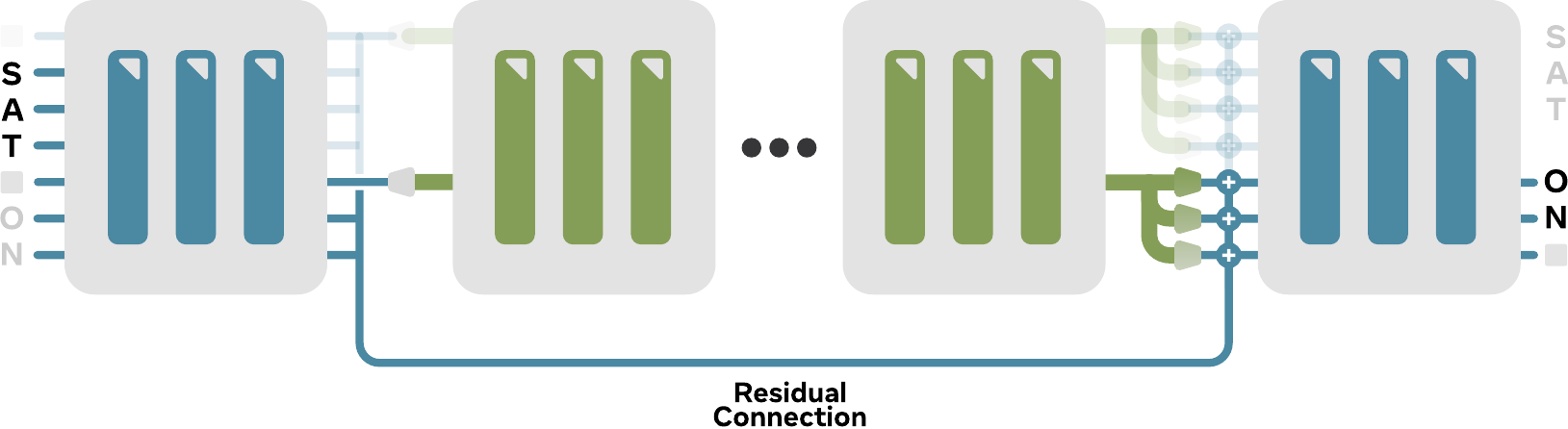}
  \caption{Pooling simply selects the vectors at the positions specified by the splitting function.
          Upsampling then expands each pooled vector to fill the next segment, applying a separate linear layer for each position. 
          For instance, the pooled vector representing the word `SAT␣' is used to help predict `ON␣'. 
          This offset lets deeper stages predict further ahead in the sequence.
          When using 4 stages, for example,  this results in the deepest stage helping for the prediction of the next four words.}
  \label{fig:poolnup}
\end{figure}

\textbf{Pooling.}  
We adopt the simplest pooling strategy: selecting the indices identified by the splitting function and projecting them to the next stage's dimensionality using a linear layer. 
Since the preceding layers already include attention mechanisms, we rely on these to do the pooling implicitly instead of relying on explicit cross attention as used in \citet{hourglass,blt}.

\textbf{Upsampling.}  
The upsampling step maps coarse representations to finer ones for the next stage. As illustrated in~\Cref{fig:poolnup}, we duplicate each coarse vector to match the length of the \textbf{following} segment, applying distinct, position-specific linear transformations to these duplicates. 
Since these transformations are shared across segments but vary by position within a segment, we term this \emph{Multi-Linear Upsampling}.
In our experiments, models with multiple stages are more sensitive to the specific choice of upsampling strategy, whereas for pooling, many strategies work equally well.

\subsubsection{Generation}\label{sec:generation}
During training, we process the entire input sequence in parallel, activating all stages simultaneously.
At inference, generation is autoregressive: the byte-level stage is active at every step, while deeper stages activate less frequently according to the pooling pattern. 
Skip connections transmit information upward at each stage, so deeper stages can integrate fine-grained details. 
This cascading, conditional activation enables efficient inference: computationally intensive high-level stages activate rarely, but still effectively guide detailed lower-level predictions.
In practice, this means that we need to cache the latest vector at the output of each stage to correctly propagate deeper stages' outputs.
\subsection{Splitting Function}\label{sec:splitting}

The \ourmodel{} architecture supports flexible splitting strategies to define pooling points at each hierarchical stage. 
The primary constraint is that any chosen splitting function must be \emph{stable to rightward insertion}: appending bytes should not alter prior pooling decisions, ensuring consistent autoregressive generation. 
Various methods (e.g., fixed windows~\citep{hourglass}, entropy~\citep{blt}, learned rules) are possible. Our current work splits on spaces using different regular expressions at each stage (details in Appendix~\ref{sec:regex}).

This strategy defines a hierarchy: Stage 1 processes raw bytes; Stage 2 pools at word boundaries (identified by the regex); Stage 3 pools after every two words(or sentence end); and Stage 4 after every four words (or sentence end). 
This rule-based approach, inspired by pre-tokenization in systems like GPT-4o's~\citep{dagan2024getting}, is effective for Latin scripts. 
Extending robustly to languages without clear delimiters remains future work. 
Unlike prior approaches~\cite{blt,hierarchicalattention,spacebyte} that used similar splits mainly to replace BPE in a single-stage context, \ourmodel{} uses these user-defined splits for its multi-stage hierarchical processing.

\subsection{Evaluating on different scales}
\label{sec:scalinglaws}

Large language models scale very predictably~\cite{kaplan,hoffmann2022training,bi2024deepseek}.
This allows us to estimate the performance of a model for a large compute budget. 
But more surprisingly, it allows us to predict the optimal hyperparameters for models way beyond our ablation budget.
\citet{bi2024deepseek} described a method for sweeping learning rates and batch sizes across a range of small models, and they demonstrated that these results can be used to predict optimal hyperparameters for larger models. 
Following their methodology, we show a different evolution of hyperparameters, both due to the data in our setup and to the hierarchical model. These hyperparameters are then used to do scaling laws for a bigger range of compute budgets to compare the baseline architecture and \ourmodel{}.
Throughout this paper, the \emph{scale} of a run is its total pre-training compute $C$ measured in Floating Point Operation (FLOP):
\[
C = 
\underbrace{F_{\text{model / input-unit}}}_{\text{FLOPs per (forward+backward) pass per input unit}} \times
\underbrace{N_\text{input-unit}}_{\text{number of units of training input}}.
\]
Following \citet{bi2024deepseek}, we define model size as the number of FLOPs per input unit instead of relying on the number of parameters.
This allows us to compare models with different architectures fairly.
The formula for the number of FLOP per input-unit for a decoder-only transformer is given by:
\[
F_{\text{model / input-unit}}
= \underbrace{6N_{\text{params}}^{\text{no-embed}}}_{\text{linear term}} + \underbrace{6d\,L\,S}_{\text{attention term}}.
\]
where, $N_{\text{params}}^{\text{no-embed}}$ is the number of parameters, excluding the embeddings. $d$ is the dimension, $S$ the sequence length and $L$ the number of layers.
To scale up, one can either make the model bigger (\(F_{\text{model / input-unit}}\uparrow\)), give it more data (\(N_{\text{input-unit}}\uparrow\)), or do both.  
\citet{gadre2024language} showed that keeping the \emph{\dm{}} $\gamma_\text{input-unit}$ constant is key to getting smooth scaling laws and predictable performance, where:
\[
\gamma_\text{input-unit}=\frac{N_\text{input-unit}}{F_{\text{model / input-unit}}}.
\]
We adopt this convention in all experiments and report the \dm{} $\gamma_\text{input-unit}$ used in the experiments.

\textbf{Bytes versus tokens.}
On DCLM, a token sequence is on average \(k\approx4.56\) times shorter than its byte sequence when using the LLaMa 3 tokenizer. 

Given some compression factor $k$ between bytes and tokens, we want to express the equivalent $\gamma_\text{bytes}$. 
To do this, we note that $N_\text{byte} = k \times N_\text{token}$ and 
$F_\text{model/byte} = F_\text{model/token} / k$. Therefore, 
\[
\gamma_\text{byte} = k^2\frac{N_\text{token}}{F_\text{model/token}} = k^2\gamma_\text{token}.
\]
This factor allows us to compare the performance of our model with the baseline on the same scale, as they will have seen the same amount of data and spent the same amount of FLOPs per token. 
Throughout the paper, we always express the \dm{} in LLaMa 3 tokens ($\gamma_\text{token}$).

\textbf{FLOPS per byte for \ourmodel{}.}
In the case of \ourmodel{}, we cannot use the same formula as the baseline because of the contraction and expansion happening in the model.
However, we can still use the same formulas as long as we account for the contraction at each stage.
So the total FLOPs per byte for \ourmodel{} is simply the sum of each stage divided by the contraction factor.
\[
F_{\text{model/byte}} = \sum_{i=1}^{L} \frac{F^{i}_{\text{model/byte}}}{k_i},
\]
where $k_i$ is the contraction factor at stage $i$.

This property allows us to have models with a higher number of parameters for the same compute budget and \dm{}.

\textbf{Hyperparameter scaling laws}
\citet{bi2024deepseek} showed that the regularity of scaling laws can be exploited to tune very large models from a sweep over much smaller ones.  
We replicate their protocol on six miniature versions of each architecture (baseline Transformer and \ourmodel{}):
we perform a quasi-random search over batch size and learning rate, keep the configurations within 1\% of the best validation loss, and fit \(\text{BSZ}(C)=A\,C^{\alpha}\) and \(\text{LR}(C)=B\,C^{\beta}\) to those points,  with parameters $A,\alpha,B\text{ and }\beta$.
We find the following formulas at the byte level for \ourmodel{}:
\[
\text{BSZ}_{\text{\ourmodel}}(C) = 0.66 C^{0.321} \qquad \text{LR}_{\text{\ourmodel}}(C) = 6.6\times C^{-0.176}.
\]
And we run the same tuning for the BPE baseline, for which we find:
\[
\text{BSZ}_{\text{BPE}}(C) = 29.9 C^{0.231} \qquad \text{LR}_{\text{BPE}}(C) = 19.3\times C^{-0.177}.
\]

\section{Experimental Results}\label{sec:experiments}
\subsection{Experimental Setup}
\textbf{Data.}  
For all experiments, we used DCLM~\citep{dclm} as our pretraining dataset, excluding a very small fraction for validation. This is around 4T training tokens (of GPTNeoXTokenizer). 
The corpus is mostly English and targets mainly natural language understanding, i.e., it contains a marginal amount of code or maths.\\
\textbf{Baselines.} 
We compare our approach to three different baselines: Transformers equipped with the BPE tokenizer of LLaMa 3, Transformers trained directly on bytes, and Mamba~\citep{mamba} trained directly on bytes. To keep the comparison fair, we trained each baseline with the same amount of data or compute. For example, if a data budget of $273$B training bytes is used to train the bytes level or \ourmodel{} model, this budget is converted to 60B training tokens for a transformer with LLaMa 3 tokenizer~\citep{llama3} because of the $4.56$ compression rate measured on the DCLM corpus.\\
\textbf{Hyperparameters.} 
For a detailed overview of the hyperparameters, see appendix~\ref{app:hypperparameters}. As explained in~\cref{sec:scalinglaws}, we sweep batch size and learning rate values across model scales ranging from 25M to 500M. Then, we extrapolate the best learning rate and batch size for any given compute budget.\\
%
%
\textbf{Evaluation Metrics.}  
All models are evaluated on a broad set of downstream tasks in a zero-shot setting, occasionally including a few in-context examples directly in the prompt. 
These tasks fall into two categories: (i) multiple-choice (MCQ) tasks, where the correct answer is selected as the option with the lowest normalized negative log-likelihood (divided by the number of characters)~\cite{brown2020language}; and (ii) open-ended generation tasks, where the model is allowed to freely generate its answer.\\
To highlight the strengths of \ourmodel{}, we include specialized benchmarks targeting character-level manipulation (CUTE~\cite{cute} \cref{app:cute}) and low-resource language translation (FLORES-200,~\cite{nllb} \cref{sec:extendedeval}). 

For clarity, we report a selection of key benchmark results in the main tables, including Hellaswag, ARC-Easy, ARC-Challenge, MMLU, NQ, TQA, and GSM8K. Also, we report 95\% confidence intervals for all tables using bootstrap.
A full breakdown of all evaluation results is provided in the appendix~\ref{app:allbenchmarks}.

In addition to task performance, the total training FLOPs and training throughput are provided for each model, measured in bytes per second per GPU (bps) on H100 80GB GPUs (internal cluster) during the actual training.

\textbf{Implementation Details.}
As scaling is key to the success of large language models, our implementation balances efficiency and simplicity. We use \emph{sequence packing} along with full attention, a strategy shown to have little to no impact on downstream performance~(\cite{dclm}). 
To reduce GPU memory pressure, all our experiments rely on Fully Sharded Data Parallelism (FSDP).

For additional speed-ups, the entire model is compiled with \texttt{torch.compile}. 
Compilation, however, requires a static computation graph, which clashes with the variable-length outputs produced by our adaptive pooling: the number of bytes per word (and thus per stage) naturally varies across sentences. 
We resolve this by fixing a maximum sequence length at every stage: sequences that exceed the limit are truncated abruptly, and shorter ones are padded. 
This compromise yields a graph that is static for compilation while still supporting adaptive hierarchical pooling in practice.
\providecommand{\err}[1]{\ensuremath{\kern0.1em{\color{gray}\scriptscriptstyle\pm#1}}}
\begin{table}[t]
    \centering    
    \scriptsize{%
    \caption{\textbf{Downstream results comparing \ourmodel{} to BPE and byte-level baselines.} 
    We report accuracy on key benchmarks with 95\% confidence intervals where applicable. 
    Literature models are shown in \textit{italics}; all models are trained on the same corpus, unless specified.
    \ourmodel{} variants differ in the number of stages. We also report compute budget and empirical training speeds in bytes/sec.\vspace{0.5em}}\label{tab:datamodel}
    \begin{tabularx}{\linewidth}{
  >{\hsize=1.0\hsize}R|
  >{\hsize=.14\hsize}C
  >{\hsize=.13\hsize}C
  >{\hsize=.13\hsize}C
  >{\hsize=.13\hsize}C|
  >{\hsize=.27\hsize}L
  >{\hsize=.27\hsize}L
  >{\hsize=.27\hsize}L
  >{\hsize=.27\hsize}L
  >{\hsize=.27\hsize}L
  >{\hsize=.27\hsize}L
  >{\hsize=.27\hsize}L}
    \toprule
        Model & Params & Emb. & Flops & bps & Hellaswag & ARC E & ARC C& MMLU & NQ & TQA & GSM8k\\ 
        \midrule
        \multicolumn{12}{l}{\textbf{Dim=2048 (1B model), 60B tokens (\dm{} of 10)}} \\
        \midrule
        Transf. Byte & 1.3B & 1M & 4e21 & 47k & 63.0~\err{1.0} & 61.2~\err{1.9} & 34.7~\err{2.7} & 24.7~\err{0.7} & 8.8~\err{0.9} & 21.4~\err{0.8} & 2.5~\err{0.9}\\
        Mamba Byte & 1.3B & 1M & 3e21 & 32k & 63.0~\err{0.9} & 60.3~\err{2.0} & 33.6~\err{2.8} & 25.1~\err{0.7} & 8.2~\err{0.9} & 21.2~\err{0.7} & 2.1~\err{0.8}\\
        Transf. BPE & 1.8B & 525M & 7e20 & 210k & 63.6~\err{1.0} & 62.8~\err{1.9} & 36.5~\err{2.7} & 26.2~\err{0.7} & 8.8~\err{0.9} & \textbf{26.3}~\err{0.8} & 2.3~\err{0.8}\\
        \ourmodel{} 2 & 1.3B & 1M & 5e20 & 225k & 64.2~\err{0.9} & 64.4~\err{1.9} & 35.2~\err{2.8} &  24.8~\err{0.7} & 8.8~\err{0.9} & 20.4~\err{0.7} & 2.7~\err{0.9}\\
        \ourmodel{} 3 & 2.5B & 1M & 7e20 & 180k & \textbf{67.4}~\err{0.9} & 65.9~\err{1.9} & 36.7~\err{2.7} & 26.3~\err{0.7} & \textbf{9.6}~\err{1.0} & 22.6~\err{0.8} & 2.3~\err{0.8}\\
        \ourmodel{} 4 &  4.2B  & 1M & 8e20 & 155k & 66.4~\err{0.9} & \textbf{67.4}~\err{1.9} & \textbf{37.0}~\err{2.8} & 26.3~\err{0.7} & 5.1~\err{0.7} & 15.5~\err{0.7} & 3.5~\err{1.0}\\
        \midrule
        \multicolumn{12}{l}{\textbf{Dim=2048 (1B model), 370B tokens (\dm{} of 40)}}\\
        \midrule
        Transf. BPE & 1.8B & 525M & 4e21 & 210k & 70.2~\err{0.9} & 68.6~\err{1.9} & 38.5~\err{2.8} & 27.0~\err{0.7} & 13.5~\err{1.1} & 37.2~\err{0.9} & 4.4~\err{1.1}\\
        \ourmodel{} 2 & 1.3B & 1M & 3e21 & 225k & 69.9~\err{0.9} & 68.6~\err{1.9} & 38.9~\err{2.7} & 28.8~\err{0.7} & 13.0~\err{1.1} & 32.5~\err{0.9} & 3.0~\err{0.9}\\
        \ourmodel{} 3 & 2.5B & 1M & 4e21 & 180k & 72.9~\err{0.9} & 72.3~\err{1.8} & \textbf{43.3}~\err{2.8} & 28.0~\err{0.7} & \textbf{15.3}~\err{1.2} & \textbf{39.1}~\err{0.9} & 3.7~\err{1.0}\\
        \ourmodel{} 4 &  4.2B  & 1M & 5e21 & 155k & \textbf{73.7}~\err{0.9} & \textbf{72.6}~\err{1.8} & 43.2~\err{2.9} & \textbf{31.7}~\err{0.7} & 14.0~\err{1.1} & 35.5~\err{0.9} & \textbf{5.3}~\err{1.2}\\
    \midrule
    \textit{DCLM-1B-5\texttimes (145B)}$^1$ & 1B  & 207M & 1e21 & - & 66.1 & 70.2 & 40.6 & 26.4 & - & 29.3 & 1.1\\
    \textit{MegaByte (263B)}$^2$ & 1.1B  & - & - & 73k & 38.9 & 54.9 & 23.4 & 25.1 & - & 9.6\\
    \textit{Hierarchical (263B)}$^3$ & 1.1B  & - & 1e21 & - & 46.5 & 65.0 & 30.5 & 26.0 & - & 9.6 & -\\
    \midrule
    \multicolumn{12}{l}{\textbf{Dim=4096 (8B model), 200B tokens (\dm{} of 5)}}\\
    \midrule
    Transf. BPE & 7.5B & 1B & 9e21 & 43k & 77.2~\err{0.8} & 74.5~\err{1.8} & \textbf{49.2}~\err{2.8} & 49.6~\err{0.8} & \textbf{21.1}~\err{1.4} & \textbf{51.1}~\err{0.9} & \textbf{10.7}~\err{1.7}\\
    \ourmodel{} 2 & 7.9B & 1M & 1e22 & 41k & \textbf{79.1}~\err{0.8} & \textbf{80.0}~\err{1.6} & \textbf{51.2}~\err{2.9} & \textbf{51.1}~\err{0.8} & \textbf{22.1}~\err{1.3} & \textbf{50.9}~\err{0.9} & \textbf{10.0}~\err{1.6}\\
    \midrule
    \textit{DCLM-7B-2\texttimes (276B)}$^1$ & 7B & 413M & 1e22 & - & 77.8 & 78.1 & \textbf{52.6} & \textbf{50.8} & - & \textbf{50.9} & 4.3\\
    \textit{Hierarchical (263B)}$^3$ & 9.2B & - & 1e22 & 15k & 56.3 & 76.6 & 44.2 & 32.0 & - & 33.1 & -\\
     \textit{BLT (220B)$^\times$}$^4$ & 8B & - & 1e22 & - & 72.2 & 66.8 & 38.8 & 25.2 & - & - & -\\
     \midrule
     \midrule
     \textit{BLT (1T)$^*$}$^4$ & 8B & - & 5e22 & - & 80.6 & 79.6 & 52.1 & 57.4 & - & - & -\\
     \textit{DCLM-7B (2.5T)$^*$}$^5$ & 7B & 413M & 1e23 & - & 80.4 & 82.2 & 59.9 & 63.7 &  - & 52.7 & 2.5\\
     \textit{LLaMa 3.1 (15T)$^\times$}$^5$ & 8B & 1B & 6e23 & - & 83.3~\err{0.8} & 80.7~\err{1.5} & 54.8~\err{2.9} & 66.4~\err{0.8} & 29.1~\err{1.5} & 64.4~\err{0.9} & 54.7~\err{2.7}\\
     \midrule
     \multicolumn{12}{l}{$^*$ Trained on mix of DCLM and other datasets}\\
     \cmidrule(lr){1-1}
     \multicolumn{12}{l}{$^\times$ Trained on different corpus than DCLM}\\
     \midrule
     \multicolumn{12}{l}{$^1$ DCLM~\cite{dclm}}\\
     \multicolumn{12}{l}{$^2$ MegaByte~\cite{yu2023megabyte}}\\
     \multicolumn{12}{l}{$^3$ Hierarchical~\cite{hierarchicalattention}}\\
     \multicolumn{12}{l}{$^4$ BLT~\cite{blt}}\\
     \multicolumn{12}{l}{$^5$ LLaMa 3.1~\cite{llama3}}\\
    \bottomrule
    \end{tabularx}
    }
\end{table}
\subsection{Equal Data Budget Results}
We evaluate the effectiveness of hierarchical pooling by fixing the model's primary hidden dimension to $2048$ and maintaining a constant total training-data budget.
The hidden dimension at each stage is scaled proportionally to its contraction ratio as described in~\cref{sec:ARUNet}. 
For instance, the byte-level stage uses a dimension of $2048 / 4 = 512$, the word-level stage uses $2048$, and the 2-word level uses $1.5 \times 2048 = 3072$, continuing in this manner for deeper stages. 
We assess the downstream performance of language models with 2, 3, and 4 stages at the 1B parameter scale. 
For the 8B model, we evaluate only the 1-stage configuration for now. 
All variants are compared against a Transformer baseline using the LLaMA 3 tokenizer of the same main hidden dimension. 
More ablations regarding pooling and the number of layers per stage can be found in the appendix~\ref{app:ablation}.

As shown in~\cref{tab:datamodel}, hierarchical models consistently match or outperform their BPE-based counterparts. This trend holds across various configurations and becomes especially pronounced as we introduce more hierarchical stages. Notably, multi-stage \ourmodel{} models (e.g., \ourmodel{}~3 and \ourmodel{}~4) outperform BPE baselines on several benchmarks. 

An interesting exception to this pattern is the TQA benchmark, which is a knowledge-intensive task evaluating the generation of the model. \ourmodel{} models along with byte-level baselines consistently underperform on TQA compared to BPE-based models. This suggests that the performance gap may not stem solely from the hierarchical structure. However, as model size and training data scale (e.g., at the 8B or 1B, 370B tokens scale), this discrepancy seems to vanish.

We observe early signs of diminishing returns beyond a certain hierarchical depth. While \ourmodel{}~4 improves on reasoning-heavy tasks such as ARC-C and GSM8k, gains on benchmarks like Hellaswag and TQA are less consistent. However, this effect may stem not from hierarchy itself, but from data efficiency: deeper hierarchies might require more training data to fully realize their potential. Supporting this interpretation, we find that \ourmodel{}~2 and \ourmodel{}~4 benefit significantly from additional training data, and that MMLU and GSM8k performances continue to improve with increased stage, even at fixed scale.

Finally, when comparing our models to similarly sized baselines from the literature (italicized in the table), we find that \ourmodel{} remains competitive, even while using significantly less training data. For instance, \textsc{BLT (1T)} uses approximately 5× more compute than our 8B model, while only being better on MMLU. Importantly, comparisons with literature models are fair, as all were trained on the same corpus: DCLM (except for \textsc{BLT (220B)} and \textsc{LLaMa 3.1 (15T)}).

To further evaluate our approach, we now turn to scaling laws to better quantify how our architecture compares to a standard Transformer with BPE. We focus on \ourmodel{}~2 and \ourmodel{}~3, using a  \dm{} of 2. This choice is motivated by the diminishing returns observed when moving from \ourmodel{}~3 to \ourmodel{}~4 under the same \dm{}.
\begin{figure}[ht]
    \centering
    \includegraphics[width=1.0\textwidth]{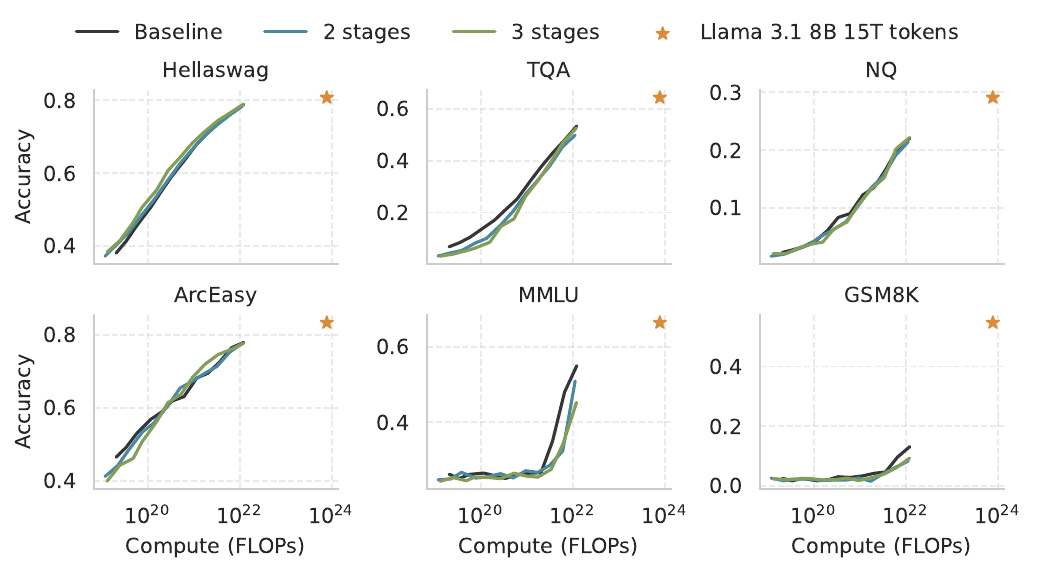}
    \caption{\small
             Downstream task performance scaling with compute (1e19-1e22 FLOPs). 
             \ourmodel{} (2/3 stages) generally tracks a strong BPE Transformer baseline, which itself performs competitively against much larger models (e.g., LLaMa 3.1 8B on 15T tokens ~ 100x compute).
             While \ourmodel{} matches the baseline on tasks like Hellaswag and ARC Easy, and catches up on TQA at higher compute, its performance improvement phase on MMLU and GSM8K appears to start later. 
             The general underperformance on GSM8K is also linked to limited math data in the DCLM pretraining corpus.}
    \label{fig:scalinglaws}
\end{figure}
\subsection{Scaling laws}
Using the learning rate and batch size formulas (\Cref{sec:scalinglaws}), we run pretrainings for a range of compute budgets ranging from 1e19 to 1e22 flops (corresponding to models from $150$M to $5.3$B non embedding parameters) for the baseline, with a \dm{} of 10. This is roughly $2\times$ the optimal \dm{} found by \citet{kaplan}. 

The list of models chosen for each budget is detailed in the appendix \ref{app:listmodel}.
\Cref{fig:scalinglaws} shows the evolution of performance on 6 downstream tasks for \ourmodel{} and the BPE baseline.
Here we mainly notice that 2 and 3 stage \ourmodel{} models can match the performance of the BPE baseline when carefully controlling for compute budget.
This is the case for Hellaswag, Arc Easy, and NQ. For TQA, \ourmodel{} both for 2 and 3 stages starts with a performance gap, but the 3 stage model catches up at 1e22 flops.
However, both 2-stage and 3-stage \ourmodel{} models are still behind the BPE baseline at 1e22 flops for GSM8K and MMLU. 
Most downstream tasks follow a sigmoid pattern: performance is near chance at low compute, then rapidly improves before plateauing.
For \ourmodel{} models, this transition appears to occur slightly later on tasks like GSM8K and MMLU, suggesting that the benefits of a deep hierarchy may become more pronounced at larger scales.
Nevertheless, on many benchmarks, both our \ourmodel{} variants and our BPE baseline achieve results remarkably close to those of considerably larger models like LLaMa 3.1 8B (pretrained on 15T tokens, representing 100 times more compute than our largest run shown here).
This proximity underscores the strength of our BPE baseline, making \ourmodel{}'s ability to match or trend towards it particularly noteworthy.
The primary exception where this close tracking is less apparent is GSM8K; however, this underperformance across all our models is likely due to the pretraining corpus, as DCLM contains very little math data.
\begin{table}[t]
  \centering
  \caption{Multilingual evaluation. \textbf{Left:} BLEU scores on the FLORES-200 benchmark across multiple languages. Higher scores indicate better translation quality. \textbf{Right:} MMLU Exact Match (\%) across 26 non-English languages. Results are averaged per language across all tasks.}
  \begin{subtable}[b]{0.535\textwidth}
    \centering
    \scriptsize{
\begin{tabularx}{\linewidth}{l|cc|cc}
  \toprule
  \multirow{2}{*}{\shortstack{\textbf{FLORES-200} \\ (BLEU)}}
    & \multicolumn{2}{c}{Lang. → Eng.} 
    & \multicolumn{2}{c}{Eng. → Lang.} \\ 
    & BPE & \ourmodel{}~2 & BPE &  \ourmodel{}~2 \\
   \midrule
  German & 34.4\err{1.2} & 33.9\err{1.2} & \textbf{16.7}\err{0.8} & 15.6\err{0.7} \\
  Dutch & 24.7\err{1.0} & 25.0\err{1.0} & \textbf{12.3}\err{0.6} & 11.7\err{0.6} \\
  Afrikaans & 32.0\err{1.3} & \textbf{35.7}\err{1.3} & 14.8\err{0.8} & \textbf{16.1}\err{0.8} \\
  Faroese & 8.7\err{0.7} & \textbf{9.9}\err{0.8} & 1.8\err{0.3} & \textbf{2.9}\err{0.4} \\
  Icelandic & 7.8\err{0.6} & \textbf{9.0}\err{0.7} & 1.7\err{0.3} & \textbf{2.5}\err{0.3} \\
  Limburgish & 15.3\err{0.9} & \textbf{19.9}\err{1.0} & 5.7\err{0.4} & \textbf{6.7}\err{0.5} \\
  Luxembourgish & 11.4\err{0.8} & \textbf{14.7}\err{0.9} & 2.6\err{0.3} & \textbf{4.0}\err{0.3} \\
  \midrule
  Italian & 29.1\err{1.0} & \textbf{30.1}\err{1.0} & 15.1\err{0.7} & 15.3\err{0.6} \\
  Friulian & 14.6\err{0.8} & \textbf{19.1}\err{1.0} & 3.2\err{0.3} & \textbf{4.0}\err{0.3} \\
  Ligurian & 16.5\err{0.9} & \textbf{21.8}\err{1.0} & 3.4\err{0.3} & 3.9\err{0.3} \\
  Lombard & 12.9\err{0.9} & \textbf{19.2}\err{1.0} & \textbf{5.2}\err{0.4} & 4.2\err{0.3} \\
  \midrule
  Sardinian & 14.3\err{0.8} & \textbf{18.2}\err{1.0} & 4.3\err{0.4} & 4.5\err{0.4} \\
  Sicilian & 11.7\err{0.8} & \textbf{16.8}\err{0.9} & 3.9\err{0.4} & \textbf{4.7}\err{0.4} \\
  Venetian & 19.8\err{1.0} & \textbf{25.4}\err{1.1} & 5.8\err{0.4} & 5.6\err{0.4} \\
  \midrule
  Spanish & 28.2\err{1.0} & \textbf{29.3}\err{1.0} & 20.2\err{0.7} & 19.8\err{0.7} \\
  Asturian & 24.0\err{1.1} & \textbf{28.6}\err{1.1} & \textbf{10.3}\err{0.6} & 8.2\err{0.5} \\
  Catalan & 28.1\err{1.1} & \textbf{33.0}\err{1.2} & 9.6\err{0.5} & \textbf{10.7}\err{0.6} \\
  Occitan & 28.0\err{1.2} & \textbf{35.5}\err{1.2} & 4.8\err{0.4} & \textbf{6.2}\err{0.4} \\
  \midrule
  Portuguese & 42.0\err{1.3} & \textbf{43.6}\err{1.3} & 25.3\err{1.0} & 25.4\err{1.0} \\
  Galician & 29.6\err{1.1} & \textbf{34.0}\err{1.2} & 9.9\err{0.5} & 10.2\err{0.6} \\
  Papiamento & 17.3\err{0.9} & \textbf{22.1}\err{1.1} & 2.5\err{0.3} & \textbf{6.3}\err{0.4} \\
  Kabuverdianu & 13.7\err{0.9} & \textbf{20.8}\err{1.1} & 2.4\err{0.3} & \textbf{5.1}\err{0.4} \\
  \midrule
  Esperanto & 15.9\err{1.0} & \textbf{19.3}\err{1.0} & 3.6\err{0.4} & \textbf{5.9}\err{0.4} \\
  \midrule
  \textbf{Average} & 20.9\err{0.2} & \textbf{24.6}\err{0.2} & 8.0\err{0.1} & \textbf{8.7}\err{0.1}\\
  \bottomrule
\end{tabularx}
}\label{tab:flores200}

    \label{fig:multilingual-flores}
  \end{subtable}%
  \hspace{0.04\textwidth}
  \begin{subtable}[b]{0.31\textwidth}
    \centering
    \scriptsize{
\begin{tabularx}{\linewidth}{lcc}
  \toprule
  \textbf{MMLU} & BPE & \ourmodel{}~2 \\ \midrule
  English   & 49.6\err{0.8} & \textbf{51.1}\err{0.8} \\ 
  \midrule
  Arabic & 29.1\err{0.8} & 29.5\err{0.8} \\
  Bengali & 27.5\err{0.7} & 27.6\err{0.8} \\
  Chinese & \textbf{33.0}\err{0.8} & 28.0\err{0.7} \\
  Czech & 30.7\err{0.8} & \textbf{32.2}\err{0.8} \\
  Dutch & 34.5\err{0.8} & \textbf{37.1}\err{0.8} \\
  Finnish & 29.0\err{0.7} & 29.3\err{0.7} \\
  French & 37.3\err{0.8} & \textbf{40.7}\err{0.8} \\
  German & 36.0\err{0.8} & \textbf{37.6}\err{0.8} \\
  Greek & 29.2\err{0.8} & \textbf{30.5}\err{0.8} \\
  Hindi & 27.9\err{0.7} & 27.5\err{0.7} \\
  Hungarian & 29.0\err{0.8} & \textbf{30.1}\err{0.8} \\
  Indonesian & 34.9\err{0.8} & \textbf{37.3}\err{0.8} \\
  Italian & 36.2\err{0.8} & \textbf{39.0}\err{0.8} \\
  Japanese & \textbf{29.5}\err{0.7} & 28.2\err{0.7} \\
  Korean & 28.4\err{0.7} & 28.2\err{0.8} \\
  Persian & 28.7\err{0.7} & 28.6\err{0.7} \\
  Polish & 30.3\err{0.8} & \textbf{32.0}\err{0.8} \\
  Portuguese & 37.2\err{0.8} & \textbf{40.9}\err{0.8} \\
  Romanian & 34.0\err{0.8} & \textbf{36.9}\err{0.8} \\
  Russian & 30.9\err{0.8} & 31.2\err{0.8} \\
  Spanish & 37.6\err{0.8} & \textbf{41.4}\err{0.8} \\
  Swahili & 28.8\err{0.7} & \textbf{29.9}\err{0.8} \\
  Swedish & 33.5\err{0.8} & \textbf{36.0}\err{0.8} \\
  Telugu & 26.8\err{0.7} & 27.4\err{0.7} \\
  Thai & 28.0\err{0.7} & 27.5\err{0.7} \\
  Turkish & 29.1\err{0.7} & \textbf{30.0}\err{0.7} \\
  Vietnamese & \textbf{31.4}\err{0.8} & 30.7\err{0.7} \\
  \midrule
  \textbf{Average} & 31.4\err{0.1} & \textbf{32.4}\err{0.1} \\
  \bottomrule
\end{tabularx}
}\label{tab:mmlu_multilingual}
    \label{fig:multilingual-mmlu}
  \end{subtable}
  \label{tab:multilingual}
\end{table}

\subsection{Extended Evaluations}\label{sec:extendedeval}
We present results highlighting two specific advantages of byte-level training with \ourmodel{} over BPE-based Transformers: improved performance on multilingual benchmarks (\Cref{tab:flores200,tab:mmlu_multilingual}) and character-level manipulation tasks (\Cref{tab:cute} in the appendix~\ref{app:cute}).

\Cref{tab:flores200,tab:mmlu_multilingual} show that both models perform surprisingly well on non-English languages, despite the fact that the training corpus (DCLM) is heavily filtered to contain mostly English.

\textbf{Cross-lingual generalization within language families.}
On the multilingual MMLU benchmark (\Cref{tab:mmlu_multilingual} right), languages using Latin scripts consistently benefit from byte-level modeling. We observe strong positive transfer between related languages. For example, Germanic languages such as German, Swedish, and Dutch show an average gain of around +3.0 points, while Romance languages like Italian, Spanish, Portuguese, and French improve by approximately +4.0 points. These results suggest that operating at the byte level allows the model to capture shared orthographic and morphological patterns across related languages.

\textbf{Transfer to low-resource languages.}
The FLORES-200 benchmark (\Cref{tab:flores200} left) includes many regional and low-resource languages that are underrepresented or absent in the training data. This setting allows us to test the model's ability to generalize based on subword morphology and shared linguistic roots. Byte-level modeling provides the flexibility to construct meaningful representations without requiring the presence of these languages in the tokenizer or training corpus. We observe consistent gains in translation tasks into English, where the model must primarily understand the source language. The advantage is particularly clear for languages that share syntactic or morphological traits with more dominant relatives in the same family. This also highlights the robustness of our model: it can produce meaningful translations even with out-of-vocabulary words or forms unseen during training. In the reverse direction (English to low-resource), generation remains more challenging.
\section{Related Work}
Traditional tokenization methods are important for computational efficiency \citep{choosetokenizer, theorytoken, retok, comptoken}, but impose fixed granularities. Early attempts to overcome this rigidity explored adaptive vocabularies \citep{adaptivetoken}, n-gram combinations \citep{tfree}, or alternative splitting criteria like entropy \citep{blt}. Our work, \ourmodel{}, advances this by integrating tokenization and representation learning into a multi-level, autoregressive U-Net architecture that operates directly on bytes.

This hierarchical, adaptive-pooling design distinguishes \ourmodel{} from prior works. For instance, Megabytes \citep{yu2023megabyte} introduce a two stage LLM using local models but with fixed-size token blocks, unlike \ourmodel{}'s input-adaptive pooling.
\citet{hierarchicalattention}, Byte Latent Transformers (BLT) \citep{blt}, and SpaceByte \citep{spacebyte} also process bytes or use specialized splitting functions. However, they typically aim to replace BPE for a single effective processing stage or use local attention mechanisms. In contrast, \ourmodel{} leverages user-defined splits within a multi-stage architecture featuring distinct pooling strategies that differ from the cross-attention methods in \citet{hourglass,blt}.
\citet{hourglass} defined a similar U-Net architecture but with fixed pooling, much smaller models, and their evaluations mainly focus on perplexity.

\section{Conclusion} 
This paper introduces \ourmodel{}, an autoregressive U-Net that processes raw bytes and learns hierarchical token representations. By dynamically pooling bytes into words and multi-word chunks, \ourmodel{} eliminates the need for predefined vocabularies and large embedding tables. Experiments show that \ourmodel{} matches strong BPE baselines under controlled compute budgets, with deeper hierarchies demonstrating promising scaling trends. Furthermore, its byte-level operation leads to improved performance on character-level tasks and better generalization to low-resource languages. This approach offers a flexible and efficient alternative to traditional tokenization methods, paving the way for more adaptable and versatile language models.
\subsection*{Limitations and further work} 
Our work uses DCLM, which is an English-only corpus. A direct limitation of our work is that it does not support non-space-based languages, and it needs a predefined splitting function. 
This shows, for example, for Chinese MMLU scores that are lower than the BPE baseline.
One extension could be to learn directly the splitting function.
On the software side, as the number of parameters increases with the number of stages, FSDP already struggles to overlap computation and communication even at 3/4 stages, it needs a minimum amount of inputs to be fully overlapped.


\bibliographystyle{unsrtnat}
\bibliography{paper}
\clearpage
\beginappendix

\section{Scaling Laws}
Every parameter goes through a multiplication and addition per input unit in the forward pass, and twice that in the backward pass resulting in 6 flops per parameter per input.
For the attention mechanism, the $QK^T$ operation dominates the computational cost, requiring $2$ FLOPs (multiply and add) per dimension for each query-key pair. 
With $d$ dimensions, $L$ layers, and a sequence length of $S$, this creates $S$ dot products per layer per input unit. 
Accounting for both forward and backward passes (3× multiplier), we get $6dLS$ FLOPs total. 
This term becomes particularly significant at smaller scales where attention costs outweight the linear parameter costs as \citet{bi2024deepseek} already point out.

Notice how the batch size is not simply a difference in constant factor but also in the exponent.
In our experiments, we find that many values of batch size and learning rates are possible and that optimal models for a given compute budget lie roughly on a line in BSZ/LR space such that both grow linearly with respect to each other.
This of course is only valid for a certain range of values above which the model becomes unstable and loses performance. 
Our hypothesis is that simply scaling the batch size such that it equals $k \times \text{BSZ}_\text{BPE}$ results in a model that is beyond that limit.

\section{Regular expression}
\label{sec:regex}

To be concrete, the regular expression used to define Stage 1 pooling is shown below:

\begin{center}
  \begin{ttfamily}\tiny
    \textcolor{blue}{( \textbackslash p\{L\}\{1,16\})}
    | \textcolor{orange}{\textbackslash p\{N\}\{1,3\}} 
    | \textcolor{teal}{ ?([\textasciicircum\textbackslash s\textbackslash p\{L\}\textbackslash p\{N\}])\{1,3\}+[\textbackslash r\textbackslash n]*}
    | \textcolor{purple}{\textbackslash s*[\textbackslash r\textbackslash n]}
    | \textcolor{red}{\textbackslash s+(?!\textbackslash S)}
    | \textcolor{gray}{\textbackslash s+}
  \end{ttfamily}
\end{center}

Each component of the regex serves a distinct role:

\begin{itemize}
    \item \textcolor{blue}{Letters (1–16 characters)}: captures typical alphabetic words.
    \item \textcolor{orange}{Numbers (1–3 digits)}: groups numerical tokens.
    \item \textcolor{teal}{Punctuation (1–3 non-alphanumeric chars)}: handles symbol groups and optional line breaks.
    \item \textcolor{purple}{Line breaks}: captures `\textbackslash r\textbackslash n` combinations and surrounding whitespace.
    \item \textcolor{red}{Trailing whitespace (non-followed by a non-space)}: captures text boundaries.
    \item \textcolor{gray}{General whitespace}: handles space separation.
\end{itemize}

\section{Ablation}\label{app:ablation}
\subsection{Pooling and Upsampling}
We describe here the different pooling and upsampling strategies explored during our experiments. While all pooling methods yielded comparable results, they offer different trade-offs in complexity and expressiveness.

\textbf{Simple Pooling.}  
This is the method used in our main experiments. We directly select the positions indicated by the splitting function and retain only those tokens.\\
\textbf{Cross-Attention Pooling.}  
A cross-attention layer is applied between the original sequence and the pooled tokens. This allows the downsampled representation to aggregate information flexibly from the full input.\\
\textbf{Average Pooling.}  
Tokens within each segment defined by the splitting function are averaged to produce a single pooled representation.\\
\textbf{Memory Layers~\cite{berges2024memory}.}  
Motivated by the concern that pooling might limit output diversity compared to embedding-table, we experimented with appending a memory layer after pooling. This layer retrieves learned embeddings based on the pooled inputs, potentially reintroducing back the diversity.

\textbf{Simple Upsampling.}  
Pooled tokens are inserted back into their original positions in the sequence, and additional context is recovered via skip connections. Earlier-layer features complement the compressed representations, and attention layers help propagate information across the sequence.\\
\textbf{Cross-Attention Upsampling.}  
A cross-attention layer is applied where each upsampled token attends to the pooled representation. This mechanism allows the model to flexibly decompress higher-level abstract representations, effectively extracting contextual information to reconstruct the outputs.\\
\textbf{Repeat Upsampling.}  
Inspired by nearest-neighbor upscaling in computer vision, each token in the compressed sequence is repeated a variable number of times, as determined by the splitting function. For this strategy to remain competitive during training, it is important to include local positional biases within each repeated segment.\\
\textbf{Multi-Linear Upsampling.}  
Each pooled token is transformed using a different linear projection for each position in the target segment. This allows upsampled tokens to vary based on their relative position while remaining conditioned on the same source. This method is used in our main experiments due to its favorable balance between simplicity and expressiveness.

\begin{table}[t]
    \centering
    \caption{Comparison between the different upsampling tools. Notice that \ourmodel{} 3 stages is much more sensitive to upsampling.}
    \label{tab:abla_pool}
    \providecommand{\err}[1]{\ensuremath{\kern0.1em{\color{gray}\scriptscriptstyle\pm#1}}}
\begin{tabularx}{\linewidth}{
  >{\hsize=1.5\hsize}R|
  *{10}{>{\hsize=.52\hsize}L}
}
\toprule
Model & Hswg & Arc\_E & Arc\_C & PIQA & SIQA & Race\_M & Race\_H & Winog & NQ & TQA \\
\midrule
\multicolumn{11}{l}{\textbf{Dim=2048 (1B model), 60B tokens (\dm{} of 10)}} \\
\midrule
\ourmodel{} 2 Simple & 62.9 ± 0.9 & 64.9 ± 1.9 & \textbf{35.5} ± 2.7 & 73.4 ± 2.0 & 45.7 ± 2.2 & 54.1 ± 2.6 & 39.3 ± 1.6 & 60.5 ± 2.7 & 7.7 ± 0.9 & 16.6 ± 0.7 \\
\midrule
\ourmodel{} 2 Average Pool & 62.5 ± 1.0 & 61.5 ± 2.0 & \textbf{35.4} ± 2.7 & 72.9 ± 2.0 & 44.7 ± 2.2 & 52.2 ± 2.6 & 36.9 ± 1.6 & 60.4 ± 2.7 & 7.2 ± 0.8 & 15.5 ± 0.7 \\
\midrule
\ourmodel{} 2 Memory Layer & 62.8 ± 0.9 & \textbf{66.5} ± 1.9 & 34.4 ± 2.7 & 72.2 ± 2.0 & 45.3 ± 2.2 & \textbf{55.2} ± 2.6 & 38.7 ± 1.6 & 61.3 ± 2.7 & 8.0 ± 0.9 & 16.6 ± 0.7 \\
\midrule
\ourmodel{} 2 Repeat Up & \textbf{64.2} ± 0.9 & 64.4 ± 1.9 & \textbf{35.2} ± 2.7 & \textbf{74.4} ± 2.0 & \textbf{46.1 }± 2.2 & 53.9 ± 2.6 & 39.0 ± 1.6 & 61.7 ± 2.7 & \textbf{8.8} ± 0.9 & \textbf{20.4} ± 0.7 \\
\midrule
\ourmodel{} 2 Multi-Linear & \textbf{63.5} ± 0.9 & 64.4 ± 1.9 & \textbf{35.3} ± 2.7 & 74.0 ± 2.0 & 45.3 ± 2.2 & 55.1 ± 2.5 & \textbf{39.6} ± 1.6 & \textbf{62.6} ± 2.6 & 8.3 ± 0.9 & 18.4 ± 0.7 \\
\midrule
\midrule
\ourmodel{} 3 Simple & 60.6 ± 1.0 & 60.8 ± 2.0 & 32.3 ± 2.7 & 72.1 ± 2.1 & 46.3 ± 2.2 & 53.1 ± 2.6 & 38.6 ± 1.6 & 62.0 ± 2.6 & 6.0 ± 0.8 & 13.3 ± 0.6 \\
\midrule
\ourmodel{} 3 Multi-Linear & \textbf{66.0} ± 0.9 & \textbf{64.1} ± 1.9 & \textbf{35.7} ± 2.7 & \textbf{75.1} ± 2.0 & \textbf{45.9} ± 2.2 & \textbf{55.4} ± 2.5 & \textbf{39.3} ± 1.6 & \textbf{64.0} ± 2.6 & \textbf{7.3} ± 0.9 & \textbf{18.7} ± 0.7 \\
\bottomrule
\end{tabularx}
\end{table}

\subsection{Layer Allocations}

To evaluate the impact of distributing different numbers of layers across stages, we conducted ablations varying the layer allocation strategy. The first stage (byte level) is fixed to three layers for all models. As shown in \cref{tab:abla_layer}, we allocate a certain percentage of the total layers to the final stage (stage 3), while ensuring that each intermediate stage retains at least three layers.

We report results for several allocation schemes, and retain the $75\%$ variant—where $75\%$ of the layers are allocated to the final stage—as the default configuration in the main paper.

\begin{table}[ht]
    \centering
    \caption{Comparison between the different percentage of layer in the last stage (the third one).}
    \label{tab:abla_layer}
    \providecommand{\err}[1]{\ensuremath{\kern0.1em{\color{gray}\scriptscriptstyle\pm#1}}}
\begin{tabularx}{\linewidth}{
  >{\hsize=1.5\hsize}R|
  *{10}{>{\hsize=.52\hsize}L}
}
\toprule
Model & Hswg & Arc\_E & Arc\_C & PIQA & SIQA & Race\_M & Race\_H & Winog & NQ & TQA \\
\midrule
\multicolumn{11}{l}{\textbf{Dim=2048 (1B model), 60B tokens (\dm{} of 10)}} \\
\midrule
AUNet 3 ($25\%$) & 65.3 ± 0.9 & 63.3 ± 1.9 & 36.0 ± 2.8 & 74.2 ± 2.0 & 46.8 ± 2.2 & 54.8 ± 2.6 & 38.7 ± 1.6 & 63.4 ± 2.6 & 8.9 ± 0.9 & 21.0 ± 0.8 \\
\midrule
AUNet 3 ($50\%$) & 66.0 ± 0.9 & 64.1 ± 1.9 & 35.7 ± 2.7 & 75.1 ± 2.0 & 45.9 ± 2.2 & 55.4 ± 2.6 & 39.3 ± 1.6 & 64.0 ± 2.7 & 7.3 ± 0.9 & 18.7 ± 0.7 \\
\midrule
AUNet 3 ($75\%$) & 67.4 ± 0.9 & 65.9 ± 1.9 & 36.7 ± 2.7 & 75.5 ± 2.0 & 46.9 ± 2.2 & 55.4 ± 2.5 & 40.5 ± 1.6 & 64.2 ± 2.6 & 9.6 ± 1.0 & 22.6 ± 0.8 \\
\bottomrule
\end{tabularx}
\end{table}

\section{Hyperparameters}\label{app:hypperparameters}
As explained in \cref{sec:scalinglaws}, we use a specific batch size and learning rate for each compite budget and architecture. Aside from this all other hyperparameters remains fixed. A summary table of all hyperparameters can be found in \cref{tab:hparams}. We use sequence packing for dataloading during training along with FSDP.
\begin{table}[ht]
    \centering
    \scriptsize{%
    \begin{tabularx}{\linewidth}{l|cccccccccc}
\toprule
Model &LR & BSZ & w.d. & lr min & grad clip & seqlen & total tokens\\
\midrule
BPE & $19.3C^{-0.177}$ & $29.9 C^{0.231}$ & 0.1 & $0.01 \times \text{lr\_max}$ & 0.2 & 2048 & $(F_{\text{model / token}})^2\gamma_\text{token}$ \\
\ourmodel{} &  $6.6C^{-0.176}$ & $0.66 C^{0.321}$ & 0.1 & $0.01 \times \text{lr\_max}$ & 0.2 & 8192 & $(F_{\text{model / byte}})^2 (20.7936\gamma_\text{token})$\\
\bottomrule
\end{tabularx}
    }
    \caption{Summary of all hyperparameters. \textit{w.d.} stands for weight decay. $\gamma_\text{tokens}$ corresponds to the \dm{} and is reported in bold in each result table, alongside the budget $C$. Flops per token/byte are detailed in table of \cref{app:listmodel}. Warmup spans $10\%$ of the total training steps, and we employ a cosine learning rate scheduler. The total number of steps is computed as $\frac{\text{total\_tokens}}{\text{BSZ}}$.}
    \label{tab:hparams}
\end{table}
\newpage

\section{CUTE Benchmark Detailed results}\label{app:cute}

\begin{wraptable}{r}{0.5\linewidth}
    \centering
    \caption{Accuracy of BPE and \ourmodel{} on word-level and letter-level tasks in CUTE.}\label{tab:cute}
\begin{tabularx}{\linewidth}{l|c|cc|cc}
  \toprule
  \multirow{2}{*}{\textbf{CUTE (EM)}} & \multirow{2}{*}{\textbf{Rand}} & \multicolumn{2}{c}{\textbf{BPE}} & \multicolumn{2}{c}{\textbf{\ourmodel{}~2}} \\  
                &      & Word & Char & Word &  Char        \\ 
  \midrule
  Spell         & 0.0  &   -  & 91.5 &   -  & \textbf{97.3} \\
  Inverse spell & 0.0  &   -  & 80.6 &   -  & \textbf{91.7} \\
  Contains      & 50.0 & \textbf{69.9} & \textbf{66.7} & 61.3 & 59.8 \\
  Delete        & 0.0  & \textbf{29.6} & 16.4 & 20.6 & \textbf{22.3} \\
  Insert        & 0.0  & \textbf{15.9} &  \textbf{9.6} &  6.5 &  7.8 \\
  Substitute    & 0.0  & \textbf{37.5} &  7.6 & 21.2 & \textbf{12.3} \\
  Swap          & 0.0  & \textbf{5.5} &  1.6 &  3.3 &  \textbf{1.9} \\
  Sem/Ortho     & 50.0 &66.0 & 40.6 & \textbf{75.1} & \textbf{48.1} \\ \midrule
  \textbf{Average}& 12.5 & \textbf{36.9} & 39.3 & 33.1 & \textbf{42.7} \\ 
  \bottomrule
\end{tabularx}
\end{wraptable}

We evaluate both the 7.5B BPE baseline and \ourmodel{}~2 on the CUTE benchmark~\cite{cute}, which tests a model's ability to manipulate both words and characters. As shown in \Cref{tab:cute}, our byte-level model performs better on character-level tasks, while the BPE baseline takes the lead on word-level ones. This reflects a natural trade-off: tokenizer-based models operate on word-like units, making them less sensitive to character structure, whereas byte-level models handle characters explicitly.

This contrast highlights a key design trade-off. Byte-level models are more flexible with unseen or morphologically rich inputs, while tokenized models benefit from stronger word-level priors. Surprisingly, despite lacking explicit character access, BPE models still perform well on spelling and reverse spelling tasks, suggesting that such skills can emerge from token-level patterns with enough capacity and data.

\section{Evaluation Benchmarks Details}\label{app:allbenchmarks}
\begin{sidewaystable}
\providecommand{\err}[1]{\ensuremath{\kern0.1em{\color{gray}\scriptscriptstyle\pm#1}}}
\begin{tabularx}{\linewidth}{
  >{\hsize=0.65\hsize}R|
  *{19}{>{\hsize=.27\hsize}L}
}
\toprule
Model & Hellaswag & Arc\_E & Arc\_C & Boolq & CSQA & MMLU & OBQA & PIQA & SIQA & Race\_M & Race\_H & Winog & NQ & TQA & GSM8k \\
\midrule
\multicolumn{16}{l}{\textbf{Dim=2048 (1B model), 60B tokens (\dm{} of 10)}} \\
\midrule
Transformer bytes & 63.0 ± 0.9 & 61.2 ± 2.0 & 34.7 ± 2.7 & 60.7 ± 1.7 & 20.0 ± 2.3 & 24.7 ± 0.7 & \textbf{38.4} ± 4.3 & 74.5 ± 2.0 & 47.1 ± 2.2 & 52.1 ± 2.5 & 37.7 ± 1.6 & 58.2 ± 2.8 & 8.8 ± 0.9 & 21.4 ± 0.8 & 2.5 ± 0.8 \\
Mamba bytes & 63.0 ± 0.9 & 60.3 ± 2.0 & 33.6 ± 2.7 & 61.2 ± 1.7 & 19.6 ± 2.3 & 25.3 ± 0.7 & \textbf{38.4} ± 4.2 & 75.1 ± 2.0 & 45.8 ± 2.2 & 46.7 ± 2.6 & 35.2 ± 1.6 & 59.1 ± 2.7 & 8.2 ± 0.9 & 21.2 ± 0.7 & 2.1 ± 0.8 \\
Transformer BPE & 63.6 ± 0.9 & 62.8 ± 2.0 & 36.5 ± 2.7 & 62.6 ± 1.6 & 18.8 ± 2.2 & 25.5 ± 0.7 & 37.4 ± 4.3 & 75.1 ± 2.0 & 45.2 ± 2.2 & 53.9 ± 2.6 & 39.3 ± 1.6 & 61.6 ± 2.7 & 8.8 ± 0.9 & 26.3 ± 0.8 & 2.3 ± 0.8 \\
AUNet 2 & 64.2 ± 0.9 & 64.4 ± 1.9 & 35.2 ± 2.7 & 62.0 ± 1.7 & \textbf{20.1} ± 2.2 & 24.5 ± 0.7 & 36.8 ± 4.2 & 74.4 ± 2.0 & 46.1 ± 2.3 & 53.9 ± 2.5 & 39.0 ± 1.6 & 61.7 ± 2.7 & 8.8 ± 0.9 & 20.4 ± 0.7 & 2.7 ± 0.9 \\
AUNet 3 & \textbf{67.4} ± 0.9 & 65.9 ± 1.9 & 36.7 ± 2.8 & 61.7 ± 1.7 & 19.7 ± 2.2 & 25.6 ± 0.7 & 38.2 ± 4.3 & \textbf{75.5} ± 2.0 & 46.9 ± 2.2 & 55.4 ± 2.5 & \textbf{40.5} ± 1.6 & \textbf{64.2} ± 2.6 & \textbf{9.6} ± 1.0 & \textbf{22.6} ± 0.8 & 2.3 ± 0.8 \\
AUNet 4 & 66.4 ± 0.9 & \textbf{67.4} ± 1.9 & \textbf{37.0} ± 2.8 & \textbf{63.3} ± 1.7 & 18.2 ± 2.1 & \textbf{25.9} ± 0.7 & 38.2 ± 4.3 & 74.5 ± 2.0 & \textbf{47.6} ± 2.3 & \textbf{55.6} ± 2.6 & 39.3 ± 1.6 & 62.0 ± 2.6 & 5.1 ± 0.7 & 15.5 ± 0.7 & \textbf{3.5} ± 1.0 \\
\midrule
\multicolumn{16}{l}{\textbf{Dim=2048 (1B model), 370B tokens (\dm{} of 40)}} \\
\midrule
Transformer BPE & 70.2 ± 0.9 & 68.6 ± 1.9 & 38.5 ± 2.8 & \textbf{62.9} ± 1.7 & 21.8 ± 2.3 & 26.3 ± 0.7 & 40.6 ± 4.3 & 76.9 ± 1.9 & 46.2 ± 2.2 & 56.7 ± 2.6 & 41.8 ± 1.6 & 65.4 ± 2.6 & 13.6 ± 1.1 & 37.2 ± 0.9 & 4.4 ± 1.1 \\
AUNet 2 & 69.9 ± 0.9 & 68.6 ± 1.9 & 38.9 ± 2.8 & 64.3 ± 1.7 & 20.8 ± 2.3 & 27.9 ± 0.7 & 39.6 ± 4.4 & 76.8 ± 1.9 & 46.6 ± 2.2 & 57.7 ± 2.5 & 42.8 ± 1.7 & 64.6 ± 2.6 & 13.0 ± 1.1 & 32.5 ± 0.9 & 3.0 ± 0.9 \\
AUNet 3 & 72.9 ± 0.9 & 72.3 ± 1.8 & \textbf{43.3} ± 2.9 & 61.8 ± 1.7 & 19.7 ± 2.3 & 27.5 ± 0.7 & \textbf{41.6} ± 4.3 & \textbf{78.1} ± 1.9 & 47.1 ± 2.3 & 58.8 ± 2.6 & 43.3 ± 1.7 & \textbf{68.7} ± 2.6 & \textbf{15.3} ± 1.2 & \textbf{39.1} ± 0.9 & 3.7 ± 1.0 \\
AUNet 4 & \textbf{73.7} ± 0.9 & \textbf{72.6} ± 1.8 & 43.2 ± 2.8 & 62.0 ± 1.7 & \textbf{23.1} ± 2.4 & \textbf{31.1} ± 0.8 & 40.8 ± 4.2 & 78.0 ± 1.9 & \textbf{47.6} ± 2.2 & \textbf{59.0} ± 2.6 & \textbf{43.9} ± 1.7 & 67.2 ± 2.6 & 14.0 ± 1.1 & 35.5 ± 0.9 & \textbf{5.3} ± 1.2 \\
\midrule
\multicolumn{16}{l}{\textbf{Dim=4096 (8B model), 200B tokens (\dm{} of 5)}} \\
\midrule
Transformer 7B & 77.3 ± 0.8 & 74.3 ± 1.8 & 49.5 ± 2.9 & 63.8 ± 1.7 & 63.2 ± 2.7 & 48.4 ± 0.8 & 43.6 ± 4.3 & 80.0 ± 1.8 & 48.4 ± 2.2 & 60.3 ± 2.5 & 43.6 ± 1.6 & 70.5 ± 2.5 & 21.1 ± 1.3 & \textbf{51.1} ± 0.9 & \textbf{10.7 }± 1.7 \\
AUNet 2 7B & \textbf{79.1} ± 0.8 & \textbf{80.0} ± 1.6 & \textbf{51.2} ± 2.9 & \textbf{68.3} ± 1.6 & \textbf{63.6} ± 2.7 & \textbf{50.0} ± 0.8 & \textbf{45.0 }± 4.4 & \textbf{80.1} ± 1.8 & \textbf{50.0} ± 2.3 &\textbf{ 61.9 }± 2.5 & \textbf{44.4} ± 1.7 & \textbf{72.2} ± 2.4 & \textbf{22.1} ± 1.3 & 50.9 ± 0.9 & 10.0 ± 1.6 \\
\midrule
\midrule
LLaMa 3.1 (15T) & 80.7 ± 0.8 & 83.3 ± 1.5 & 54.8 ± 2.8 & 75.0 ± 1.5 & 74.6 ± 2.5 & 65.4 ± 0.8 & 45.4 ± 4.4 & 80.8 ± 1.8 & 49.6 ± 2.3 & 65.3 ± 2.5 & 49.4 ± 1.7 & 74.5 ± 2.4 & 29.1 ± 1.5 & 64.4 ± 0.9 & 54.7 ± 2.7 \\
\bottomrule
\end{tabularx}
\caption{Performance of \ourmodel{} on many benchmarks.}\label{tab:landscape}
\end{sidewaystable}

\section{List of Models}\label{app:listmodel}

\section{Model Configuration Tables}

This appendix provides detailed configuration parameters for all models used in the experiments, organized into three categories for clarity.

\begin{table}[htbp]
\centering
\caption{Model architecture parameters including dimensions, layers, and FFN sizes. Semicolons separated values for different stages in hierarchical models.}
\resizebox{0.8\linewidth}{!}{%
\begin{tabular}{lcccc}
\toprule
Name & Dim & Layers & Head Dim & FFN Dim \\
\midrule
Transformer bytes 1B & 2048 & 25 & 128 & 5632 \\
Mamba bytes 1B & 2048 & 50 & 64 & 5632 \\
Transformer 1B BPE & 2048 & 25 & 128 & 5632 \\
AUNet 2 1B & 512; 2048 & 3; 25 & 64; 128 & 1536; 5632 \\
AUNet 3 1B & 512; 2048; 3072 & 3; 3; 18 & 64; 128; 128 & 1536; 5632; 8192 \\
AUNet 4 1B & 512; 2048; 3072; 4608 & 3; 3; 4; 10 & 64; 128; 128; 128 & 1536; 5632; 8192; 12288 \\
Transformer 1B dm8 BPE & 2048 & 25 & 128 & 5632 \\
AUNet 2 1B dm8 & 512; 2048 & 3; 25 & 64; 128 & 1536; 5632 \\
AUNet 3 1B dm8 & 512; 2048; 3072 & 3; 3; 18 & 64; 128; 128 & 1536; 5632; 8192 \\
AUNet 4 1B dm8 & 512; 2048; 3072; 4608 & 3; 3; 3; 12 & 64; 128; 128; 128 & 1536; 5632; 8192; 12288 \\
Transformer 7B dm1 & 4096 & 32 & 128 & 11008 \\
AUNet 2 7B dm1 & 1024; 4096 & 3; 32 & 64; 128 & 4096; 14336 \\
Scaling baseline 1e19 & 1024 & 12 & 128 & 2816 \\
Scaling baseline 2e19 & 1152 & 13 & 128 & 3072 \\
Scaling baseline 4e19 & 1280 & 14 & 128 & 3584 \\
Scaling baseline 8e19 & 1536 & 15 & 128 & 4096 \\
Scaling baseline 1e20 & 1664 & 17 & 128 & 4608 \\
Scaling baseline 3e20 & 1792 & 20 & 128 & 4864 \\
Scaling baseline 5e20 & 2048 & 21 & 128 & 5632 \\
Scaling baseline 1e21 & 2304 & 24 & 128 & 6144 \\
Scaling baseline 2e21 & 2560 & 26 & 128 & 6912 \\
Scaling baseline 3e21 & 2816 & 29 & 128 & 7680 \\
Scaling baseline 6e21 & 3072 & 34 & 128 & 8192 \\
Scaling baseline 1e22 & 3456 & 37 & 128 & 9216 \\
Scaling AUNet 2 1e19 & 256; 1024 & 3; 11 & 64; 128 & 768; 2816 \\
Scaling AUNet 2 2e19 & 256; 1152 & 3; 13 & 64; 128 & 768; 3072 \\
Scaling AUNet 2 4e19 & 256; 1280 & 3; 14 & 64; 128 & 768; 3584 \\
Scaling AUNet 2 8e19 & 384; 1536 & 3; 14 & 64; 128 & 1024; 4096 \\
Scaling AUNet 2 1e20 & 384; 1536 & 3; 19 & 64; 128 & 1024; 4096 \\
Scaling AUNet 2 3e20 & 512; 1920 & 3; 17 & 64; 128 & 1536; 5120 \\
Scaling AUNet 2 5e20 & 512; 2048 & 3; 21 & 64; 128 & 1536; 5632 \\
Scaling AUNet 2 9e20 & 512; 2304 & 3; 24 & 64; 128 & 1536; 6144 \\
Scaling AUNet 2 2e21 & 640; 2560 & 3; 26 & 64; 128 & 1792; 6912 \\
Scaling AUNet 2 3e21 & 640; 2688 & 3; 33 & 64; 128 & 1792; 7168 \\
Scaling AUNet 2 6e21 & 768; 3200 & 3; 32 & 64; 128 & 2048; 8704 \\
Scaling AUNet 2 1e22 & 896; 3584 & 3; 35 & 64; 128 & 2560; 9728 \\
Scaling AUNet 3 1e19 & 256; 1024; 1536 & 3; 3; 4 & 64; 128; 128 & 768; 2816; 4096 \\
Scaling AUNet 3 2e19 & 256; 1152; 1792 & 3; 3; 5 & 64; 128; 128 & 768; 3072; 4864 \\
Scaling AUNet 3 5e19 & 256; 1280; 1920 & 3; 3; 7 & 64; 128; 128 & 768; 3584; 5120 \\
Scaling AUNet 3 7e19 & 256; 1280; 1920 & 3; 3; 10 & 64; 128; 128 & 768; 3584; 5120 \\
Scaling AUNet 3 2e20 & 384; 1536; 2304 & 3; 3; 10 & 64; 128; 128 & 1024; 4096; 6144 \\
Scaling AUNet 3 3e20 & 384; 1536; 2304 & 3; 3; 15 & 64; 128; 128 & 1024; 4096; 6144 \\
Scaling AUNet 3 5e20 & 512; 1920; 2816 & 3; 3; 13 & 64; 128; 128 & 1536; 5120; 7680 \\
Scaling AUNet 3 9e20 & 512; 2048; 3072 & 3; 3; 16 & 64; 128; 128 & 1536; 5632; 8192 \\
Scaling AUNet 3 2e21 & 512; 2304; 3456 & 3; 3; 18 & 64; 128; 128 & 1536; 6144; 9216 \\
Scaling AUNet 3 3e21 & 640; 2560; 3840 & 3; 3; 21 & 64; 128; 128 & 1792; 6912; 10240 \\
Scaling AUNet 3 6e21 & 640; 2688; 4096 & 3; 3; 26 & 64; 128; 128 & 1792; 7168; 11008 \\
Scaling AUNet 3 1e22 & 768; 3200; 4864 & 3; 3; 26 & 64; 128; 128 & 2048; 8704; 13056 \\
\bottomrule
\end{tabular}
}
\label{tab:model_arch}
\end{table}

\begin{table}[htbp]
\centering
\caption{Training configuration including computational costs, steps, batch sizes, and tokenization}
\resizebox{\linewidth}{!}{%
\begin{tabular}{lccccccccc}
\toprule
Name & Total FLOPs & Tokens/Step & FLOPs/Token & Steps & G/Acc & Batch Size & Seq Len & NGpus & Tokenizer \\
\midrule
Transformer bytes 1B & nan & nan & nan & 60000 & 1 & 4 & 4096 & NaN & bytes \\
Mamba bytes 1B & nan & nan & nan & 60000 & 1 & 4 & 4096 & NaN & bytes \\
Transformer 1B BPE & 6.6e20 & 1.0e06 & 1.1e10 & 60000 & 1 & 4 & 4096 & 64 & tiktoken \\
AUNet 2 1B & 5.1e20 & 1.6e06 & 1.8e09 & 180000 & 1 & 12 & 8192 & 16 & bytes \\
AUNet 3 1B & 6.7e20 & 2.8e06 & 2.3e09 & 105000 & 1 & 14 & 8192 & 24 & bytes \\
AUNet 4 1B & 8.0e20 & 2.8e06 & 2.8e09 & 105000 & 1 & 14 & 8192 & 24 & bytes \\
Transformer 1B dm8 BPE & 3.6e21 & 1.2e06 & 9.9e09 & 310000 & 1 & 9 & 2048 & 64 & tiktoken \\
AUNet 2 1B dm8 & 3.2e21 & 1.8e06 & 1.8e09 & 950000 & 1 & 7 & 8192 & 32 & bytes \\
AUNet 3 1B dm8 & 4.0e21 & 5.8e06 & 2.3e09 & 300000 & 1 & 11 & 8192 & 64 & bytes \\
AUNet 4 1B dm8 & 5.0e21 & 5.8e06 & 2.9e09 & 300000 & 1 & 11 & 8192 & 64 & bytes \\
Transformer 7B dm1 & 9.5e21 & 2.1e06 & 4.5e10 & 100000 & 1 & 2 & 4096 & 256 & tiktoken \\
AUNet 2 7B dm1 & 1.2e22 & 4.2e06 & 1.0e10 & 277834 & 1 & 1 & 8192 & 128 & bytes \\
Scaling baseline 1e19 & 2.0e19 & 7.7e05 & 1.9e09 & 14008 & 15 & 25 & 2048 & 1 & tiktoken \\
Scaling baseline 2e19 & 3.3e19 & 8.8e05 & 2.3e09 & 16120 & 18 & 24 & 2048 & 1 & tiktoken \\
Scaling baseline 4e19 & 5.6e19 & 9.9e05 & 2.9e09 & 19438 & 22 & 22 & 2048 & 1 & tiktoken \\
Scaling baseline 8e19 & 1.1e20 & 1.2e06 & 4.0e09 & 24252 & 15 & 19 & 2048 & 2 & tiktoken \\
Scaling baseline 1e20 & 2.0e20 & 1.4e06 & 5.1e09 & 28160 & 13 & 17 & 2048 & 3 & tiktoken \\
Scaling baseline 3e20 & 3.3e20 & 1.6e06 & 6.5e09 & 32466 & 11 & 14 & 2048 & 5 & tiktoken \\
Scaling baseline 5e20 & 6.0e20 & 1.8e06 & 8.6e09 & 39573 & 12 & 12 & 2048 & 6 & tiktoken \\
Scaling baseline 1e21 & 1.1e21 & 2.1e06 & 1.2e10 & 46980 & 4 & 8 & 2048 & 32 & tiktoken \\
Scaling baseline 2e21 & 2.0e21 & 2.4e06 & 1.5e10 & 55900 & 4 & 9 & 2048 & 32 & tiktoken \\
Scaling baseline 3e21 & 3.6e21 & 2.8e06 & 2.0e10 & 64711 & 3 & 7 & 2048 & 64 & tiktoken \\
Scaling baseline 6e21 & 6.5e21 & 3.1e06 & 2.7e10 & 77520 & 4 & 6 & 2048 & 64 & tiktoken \\
Scaling baseline 1e22 & 1.2e22 & 3.9e06 & 3.6e10 & 84915 & 3 & 5 & 2048 & 128 & tiktoken \\
Scaling AUNet 2 1e19 & 1.1e19 & 8.7e05 & 2.4e08 & 56240 & 2 & 53 & 8192 & 1 & bytes \\
Scaling AUNet 2 2e19 & 2.2e19 & 1.1e06 & 3.2e08 & 63457 & 3 & 43 & 8192 & 1 & bytes \\
Scaling AUNet 2 4e19 & 3.7e19 & 1.3e06 & 4.2e08 & 68794 & 4 & 39 & 8192 & 1 & bytes \\
Scaling AUNet 2 8e19 & 7.6e19 & 1.6e06 & 6.0e08 & 79862 & 6 & 32 & 8192 & 1 & bytes \\
Scaling AUNet 2 1e20 & 1.3e20 & 1.8e06 & 7.9e08 & 89768 & 4 & 28 & 8192 & 2 & bytes \\
Scaling AUNet 2 3e20 & 2.6e20 & 2.4e06 & 1.1e09 & 99036 & 4 & 24 & 8192 & 3 & bytes \\
Scaling AUNet 2 5e20 & 5.0e20 & 2.9e06 & 1.5e09 & 108807 & 4 & 18 & 8192 & 5 & bytes \\
Scaling AUNet 2 9e20 & 9.3e20 & 3.7e06 & 2.1e09 & 119909 & 1 & 14 & 8192 & 32 & bytes \\
Scaling AUNet 2 2e21 & 1.7e21 & 4.2e06 & 2.9e09 & 141463 & 2 & 8 & 8192 & 32 & bytes \\
Scaling AUNet 2 3e21 & 3.1e21 & 5.2e06 & 3.9e09 & 153594 & 2 & 10 & 8192 & 32 & bytes \\
Scaling AUNet 2 6e21 & 5.9e21 & 6.3e06 & 5.3e09 & 176692 & 1 & 8 & 8192 & 96 & bytes \\
Scaling AUNet 2 1e22 & 1.1e22 & 7.9e06 & 7.3e09 & 193096 & 1 & 6 & 8192 & 160 & bytes \\
Scaling AUNet 3 1e19 & 1.3e19 & 9.0e05 & 2.5e08 & 56992 & 2 & 55 & 8192 & 1 & bytes \\
Scaling AUNet 3 2e19 & 2.4e19 & 1.1e06 & 3.4e08 & 64347 & 3 & 45 & 8192 & 1 & bytes \\
Scaling AUNet 3 5e19 & 4.7e19 & 1.4e06 & 4.8e08 & 71962 & 4 & 42 & 8192 & 1 & bytes \\
Scaling AUNet 3 7e19 & 7.3e19 & 1.6e06 & 5.9e08 & 79222 & 5 & 38 & 8192 & 1 & bytes \\
Scaling AUNet 3 2e20 & 1.5e20 & 2.0e06 & 8.6e08 & 90845 & 4 & 30 & 8192 & 2 & bytes \\
Scaling AUNet 3 3e20 & 2.7e20 & 2.4e06 & 1.1e09 & 100198 & 4 & 24 & 8192 & 3 & bytes \\
Scaling AUNet 3 5e20 & 5.2e20 & 2.9e06 & 1.6e09 & 113583 & 2 & 22 & 8192 & 8 & bytes \\
Scaling AUNet 3 9e20 & 9.2e20 & 3.7e06 & 2.1e09 & 119374 & 1 & 14 & 8192 & 32 & bytes \\
Scaling AUNet 3 2e21 & 1.7e21 & 4.2e06 & 2.9e09 & 141594 & 1 & 16 & 8192 & 32 & bytes \\
Scaling AUNet 3 3e21 & 3.3e21 & 5.2e06 & 4.0e09 & 158941 & 1 & 10 & 8192 & 64 & bytes \\
Scaling AUNet 3 6e21 & 6.0e21 & 6.3e06 & 5.4e09 & 177921 & 1 & 8 & 8192 & 96 & bytes \\
Scaling AUNet 3 1e22 & 1.2e22 & 8.4e06 & 7.6e09 & 187518 & 1 & 4 & 8192 & 256 & bytes \\
\bottomrule
\end{tabular}
}
\label{tab:training_config}
\end{table}

\begin{table}[htbp]
\centering
\caption{Optimization hyperparameters including learning rates, weight decay, and scheduler settings.}
\resizebox{0.83\linewidth}{!}{%
\begin{tabular}{lcccccc}
\toprule
Name & LR & WD & $\beta_1$ & $\beta_2$ & Scheduler & Warmup \\
\midrule
Transformer bytes 1B & 0.003 & 0.033 & 0.9 & 0.95 & cosine & 5000 \\
Mamba bytes 1B & 0.003 & 0.033 & 0.9 & 0.95 & cosine & 5000 \\
Transformer 1B BPE & 0.003 & 0.033 & 0.9 & 0.95 & cosine & 5000 \\
AUNet 2 1B & 0.00165 & 0.1 & 0.9 & 0.95 & cosine & 10000 \\
AUNet 3 1B & 0.0015 & 0.1 & 0.9 & 0.95 & cosine & 10000 \\
AUNet 4 1B & 0.0015 & 0.1 & 0.9 & 0.95 & cosine & 20000 \\
Transformer 1B dm8 BPE & 0.001 & 0.1 & 0.9 & 0.95 & cosine & 2000 \\
AUNet 2 1B dm8 & 0.00094 & 0.1 & 0.9 & 0.95 & cosine & 10000 \\
AUNet 3 1B dm8 & 0.0011 & 0.1 & 0.9 & 0.95 & cosine & 10000 \\
AUNet 4 1B dm8 & 0.0011 & 0.1 & 0.9 & 0.95 & cosine & 10000 \\
Transformer 7B dm1 & 0.001 & 0.05 & 0.9 & 0.95 & cosine & 10000 \\
AUNet 2 7B dm1 & 0.000818 & 0.1 & 0.9 & 0.95 & cosine & 5000 \\
Scaling baseline 1e19 & 0.008152 & 0.1 & 0.9 & 0.95 & cosine & 2000 \\
Scaling baseline 2e19 & 0.007378 & 0.1 & 0.9 & 0.95 & cosine & 2000 \\
Scaling baseline 4e19 & 0.006633 & 0.1 & 0.9 & 0.95 & cosine & 2000 \\
Scaling baseline 8e19 & 0.005788 & 0.1 & 0.9 & 0.95 & cosine & 2000 \\
Scaling baseline 1e20 & 0.005204 & 0.1 & 0.9 & 0.95 & cosine & 2000 \\
Scaling baseline 3e20 & 0.004693 & 0.1 & 0.9 & 0.95 & cosine & 2000 \\
Scaling baseline 5e20 & 0.0042 & 0.1 & 0.9 & 0.95 & cosine & 2000 \\
Scaling baseline 1e21 & 0.003722 & 0.1 & 0.9 & 0.95 & cosine & 2000 \\
Scaling baseline 2e21 & 0.003357 & 0.1 & 0.9 & 0.95 & cosine & 2000 \\
Scaling baseline 3e21 & 0.003018 & 0.1 & 0.9 & 0.95 & cosine & 2000 \\
Scaling baseline 6e21 & 0.002701 & 0.1 & 0.9 & 0.95 & cosine & 2000 \\
Scaling baseline 1e22 & 0.002416 & 0.1 & 0.9 & 0.95 & cosine & 2000 \\
Scaling AUNet 2 1e19 & 0.002923 & 0.1 & 0.9 & 0.95 & cosine & 10000 \\
Scaling AUNet 2 2e19 & 0.002615 & 0.1 & 0.9 & 0.95 & cosine & 10000 \\
Scaling AUNet 2 4e19 & 0.002377 & 0.1 & 0.9 & 0.95 & cosine & 10000 \\
Scaling AUNet 2 8e19 & 0.002096 & 0.1 & 0.9 & 0.95 & cosine & 10000 \\
Scaling AUNet 2 1e20 & 0.001906 & 0.1 & 0.9 & 0.95 & cosine & 10000 \\
Scaling AUNet 2 3e20 & 0.001685 & 0.1 & 0.9 & 0.95 & cosine & 10000 \\
Scaling AUNet 2 5e20 & 0.001507 & 0.1 & 0.9 & 0.95 & cosine & 10000 \\
Scaling AUNet 2 9e20 & 0.001348 & 0.1 & 0.9 & 0.95 & cosine & 10000 \\
Scaling AUNet 2 2e21 & 0.001214 & 0.1 & 0.9 & 0.95 & cosine & 10000 \\
Scaling AUNet 2 3e21 & 0.00109 & 0.1 & 0.9 & 0.95 & cosine & 10000 \\
Scaling AUNet 2 6e21 & 0.0009731 & 0.1 & 0.9 & 0.95 & cosine & 10000 \\
Scaling AUNet 2 1e22 & 0.0008719 & 0.1 & 0.9 & 0.95 & cosine & 10000 \\
Scaling AUNet 3 1e19 & 0.002872 & 0.1 & 0.9 & 0.95 & cosine & 10000 \\
Scaling AUNet 3 2e19 & 0.002561 & 0.1 & 0.9 & 0.95 & cosine & 10000 \\
Scaling AUNet 3 5e19 & 0.002279 & 0.1 & 0.9 & 0.95 & cosine & 10000 \\
Scaling AUNet 3 7e19 & 0.00211 & 0.1 & 0.9 & 0.95 & cosine & 10000 \\
Scaling AUNet 3 2e20 & 0.001852 & 0.1 & 0.9 & 0.95 & cosine & 10000 \\
Scaling AUNet 3 3e20 & 0.001678 & 0.1 & 0.9 & 0.95 & cosine & 10000 \\
Scaling AUNet 3 5e20 & 0.001496 & 0.1 & 0.9 & 0.95 & cosine & 10000 \\
Scaling AUNet 3 9e20 & 0.001351 & 0.1 & 0.9 & 0.95 & cosine & 10000 \\
Scaling AUNet 3 2e21 & 0.001213 & 0.1 & 0.9 & 0.95 & cosine & 10000 \\
Scaling AUNet 3 3e21 & 0.001077 & 0.1 & 0.9 & 0.95 & cosine & 10000 \\
Scaling AUNet 3 6e21 & 0.0009707 & 0.1 & 0.9 & 0.95 & cosine & 10000 \\
Scaling AUNet 3 1e22 & 0.0008612 & 0.1 & 0.9 & 0.95 & cosine & 10000 \\
\bottomrule
\end{tabular}
}
\label{tab:optim_params}
\end{table}

\end{document}